\documentclass[letterpaper, 10 pt, conference]{ieeeconf}  %

\IEEEoverridecommandlockouts                              %

\usepackage{graphics} %
\usepackage{epsfig} %
\usepackage{mathptmx} %
\usepackage{times} %
\usepackage{amsmath} %
\usepackage{amssymb}  %
\usepackage{graphicx}
\usepackage{cuted}
\usepackage[font=footnotesize]{caption}
\usepackage[
    backend=biber, 
    style=numeric-comp, 
    sortcites=true, 
    sorting=none, 
    hyperref=true]{biblatex}
\usepackage{xcolor}
\definecolor{MidnightBlue}{RGB}{0,103,149} %
\usepackage[colorlinks=true,
            linkcolor=black,
            citecolor=MidnightBlue,
            urlcolor=MidnightBlue]{hyperref}
\usepackage{caption}

\title{\LARGE \bf
EgoThumb: A Vision-Driven Supernumerary Robotic Thumb \\
for Autonomous Grasp and Holding Assistance
}

\author{Jianyu Hou$^{1*}$, Jiawei Fu$^{1*}$, Weikang Wan$^{1}$, Minghui Jiang$^{1}$, Jiajun Li$^{1}$, Hao Su$^{2}$ \thanks{$^{*}$These authors contributed equally.}%
\thanks{$^{1}$The authors are with the Department of Computer Science and Engineering, University of California San Diego, La Jolla, CA 92093, USA. {\tt\small \{jeh089, jif017, w2wan, mijiang, jil186\}@ucsd.edu}}%
\thanks{$^{2}$Hao Su is with Sudo AI GmbH, 8050 Zurich, Switzerland. {\tt\small academic@haosu.ai}}%
}

\begin{document}

\maketitle
\thispagestyle{empty}
\pagestyle{empty}

\begin{strip}
\vspace{-7em} %
\centering
\includegraphics[width=\textwidth]{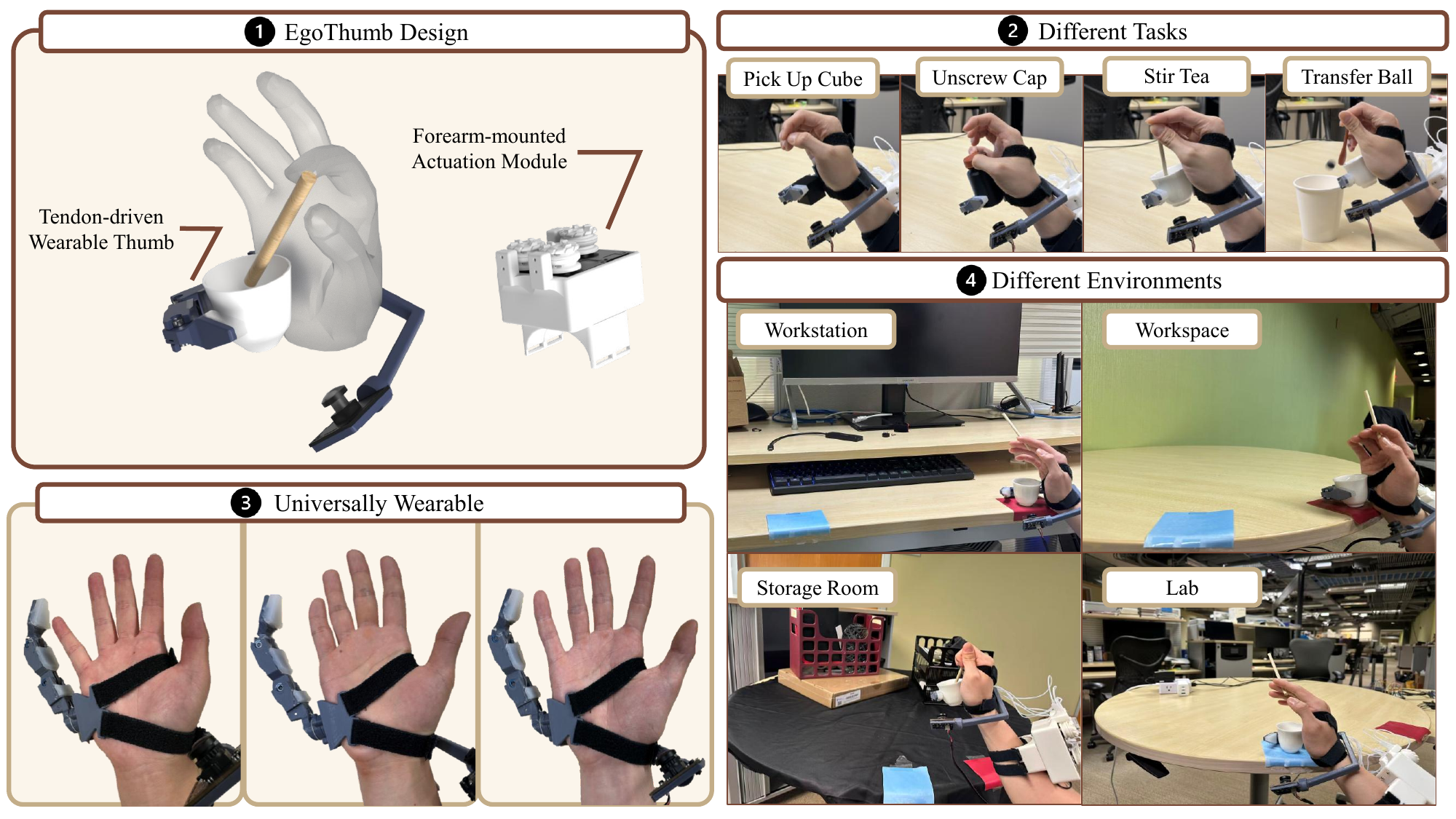}
\vspace{-1.5em} %
\captionof{figure}{
\textbf{Overview of our EgoThumb System.}
\textbf{(1) EgoThumb Design}: We present a fully open-source, vision-driven supernumerary robotic thumb, featuring a compact, tendon-driven mechanism with forearm-mounted actuation.
\textbf{(2) Different Tasks}: EgoThumb enables a single hand to independently execute complex tasks that typically require bimanual coordination.
\textbf{(3) Universally Wearable}: The device is secured via adjustable hook-and-loop straps, ensuring seamless physical adaptation across diverse hand morphologies.
\textbf{(4) Different Environments}: The system achieves strong generalization performance across varied, unstructured environments.
}
\label{fig:teaser}
\vspace{-1em} %
\end{strip}

\begin{abstract}
Supernumerary robotic fingers augment human dexterity by enabling a single hand to perform complex ``hold-and-manipulate'' tasks that typically require bimanual coordination. 
However, existing devices are often hindered by bulky hardware that limits interaction to large objects and rely on explicit hand motion triggers that lack environmental context.
To address these challenges, we present EgoThumb, a fully open-source, vision-driven supernumerary robotic thumb. 
EgoThumb features a compact, dual tendon-driven finger whose actuation is relocated to the forearm, reducing distal weight while enabling precise active joint control.
The system utilizes a visual imitation learning policy that maps egocentric images from a palm-facing camera directly to continuous joint positions.
By implicitly decoding human intent from visual cues capturing both the hand and the object of interest, EgoThumb achieves robust autonomous manipulation and strong generalization across diverse users and environments. 
All hardware designs and software implementations are fully open-sourced at \url{https://ego-thumb.github.io}.
\end{abstract}

\section{Introduction}
\label{sec:intro}
The enhancement of human physical capabilities through robotic systems has long been a central goal of robotics research.
Existing literature in this domain typically falls into three branches: exoskeletons, prostheses, and supernumerary limbs.
Exoskeletons attach motors to human joints to improve existing capabilities, such as playing instruments \cite{furuya2025surmounting} and walking \cite{chen2025biarticular}.
Prosthetic devices replace missing limbs to restore lost functions, including hand manipulation \cite{greene2024evaluating} and locomotion \cite{maldonado2025transfer}.
In contrast, supernumerary robotic (SR) limbs attach additional robotic arms and grippers to the human body to augment functional capacity \cite{ parietti2016supernumerary}.
For instance, body-mounted robotic arms can support heavy loads during overhead assembly tasks \cite{parietti2014bracing}.

While much of this prior work focuses on proximal limb augmentation, there is growing interest in mounting robotic fingers directly onto the hand or forearm to extend the degrees of freedom of the human hand \cite{wu2014supernumerary}.
These additional digits can actively collaborate with biological fingers, enabling a single hand to perform tasks that typically require bimanual coordination.
They have been explored across contexts ranging from assistive and therapeutic use for motor impairments \cite{ort2015supernumerary} to able-bodied augmentation for skilled tasks \cite{cunningham2018supernumerary}.
Specifically, we focus on ``hold-and-manipulate'' tasks \cite{wu2015hold}, where the SR finger must: 1) autonomously decide when to grasp and release an object, and 2) stably hold the object while the human hand interacts with it.

Despite their potential, existing SR fingers face two key limitations. First, their bulky hardware \cite{wu2014bio, ort2015supernumerary, fasoulas2024mechatronic} suits large objects but hinders interaction with the smaller, everyday objects central to daily manipulation, such as the teacup in Fig. \ref{fig:teaser}.2 (Stir Tea). Second, they are typically driven by teleoperation \cite{kieliba2021robotic, clode2024evaluating} or low-dimensional hand-motion triggers \cite{wu2014bio, wu2015hold}, which lack the autonomy and environmental awareness needed to infer intent from context, such as grasping the bottle in Fig. \ref{fig:teaser}.2 (Unscrew Cap) as the hand approaches, without an explicit trigger motion.

To address these limitations, we present \textbf{EgoThumb}, a fully open-source, vision-driven supernumerary robotic thumb system.
EgoThumb actively perceives hand motion and the environment through vision to autonomously determine the optimal timing for grasping, holding, and releasing objects.
The key contributions of our work are as follows:
\begin{itemize}
    \item \textbf{Hardware Design}: Our system utilizes a dual-tendon actuation strategy for each joint, enabling active control of both flexion and extension to ensure precise position control.
    Thus, the actuators are located on the forearm, resulting in a lightweight, wearable finger design.
    \item \textbf{Egocentric Perception}: To achieve autonomous control with environmental awareness, we integrate a palm-facing camera that captures the egocentric visual context of both the hand motion and the surroundings.
    \item \textbf{Intelligent Autonomy}: We formulate the control problem as a visual imitation learning task, training a policy to map egocentric images directly to joint positions.
\end{itemize}

Through these hardware and algorithmic advancements, EgoThumb achieves robust performance across four distinct real-world tasks without explicitly measuring hand motion.
It also demonstrates strong generalization capabilities across different users and varied environmental backgrounds.

\section{Related Work}
\label{sec:related-works}
\subsection{Design of Supernumerary Robotic Fingers}
While direct-driven mechanisms provide the high torque needed for large-scale supernumerary robotic (SR) limbs \cite{parietti2016supernumerary, parietti2014bracing}, applying this architecture to SR fingers \cite{wu2014bio, ort2015supernumerary, cunningham2018supernumerary, wu2016implicit, fasoulas2024mechatronic} results in bulky devices. 
This bulk limits interaction to large objects and causes user fatigue due to significant distal mass.
Similarly, pneumatic designs suffer from substantial volumetric overhead \cite{liang2019fabric}. 
Although compact tendon-driven alternatives exist \cite{kieliba2021robotic, clode2024evaluating}, they remain predominantly closed-source and rely on passive, spring-based extension, which degrades joint positioning accuracy. 
In contrast, EgoThumb is an open-source, compact SR finger whose dual-tendon drive actively controls flexion and extension for precise joint positioning. Placing the actuators on the forearm decouples the finger dimensions from motor size, unlike joint-mounted designs, minimizing distal weight and enabling a compact form suited to everyday objects.

\subsection{Control of Supernumerary Robotic Fingers}
Existing control strategies for SR fingers largely rely on teleoperation \cite{kieliba2021robotic, clode2024evaluating, cunningham2018supernumerary, liang2019fabric, liang2017design, weijun2024kinematic, katmah2023cognitive}, which inherently lacks autonomy.
Alternatively, some systems guide finger motion by mapping explicitly measured, low-dimensional hand kinematics \cite{wu2014bio, ort2015supernumerary, wu2015hold, wu2016implicit}. 
However, these reactive approaches are insufficient when grasping must be initiated based on environmental context rather than explicit hand modulation. 
In contrast, EgoThumb infers grasp timing from visual context via imitation learning, rather than relying on teleoperation or explicit hand-motion triggers.

\subsection{Imitation Learning for Robot Hands}
Imitation learning is a promising paradigm for learning robot manipulation policies from expert demonstrations. 
Recently, wearable data collection systems, such as UMI~\cite{chi2024universal}, DexUMI~\cite{xu2025dexumi}, and DexWild~\cite{tao2025dexwild}, have highlighted the efficacy of end-to-end vision-motor policies for dexterous manipulation by leveraging diverse human data. 
However, these frameworks target standalone robots, overlooking the physical human-robot interaction inherent to supernumerary devices.
EgoThumb instead frames wearable-finger control as visual imitation learning in a shared workspace, directly mapping egocentric images to joint positions.

\section{System Design}
\label{sec:method}
\begin{figure*}[!t]
  \centering
  \includegraphics[width=\textwidth]{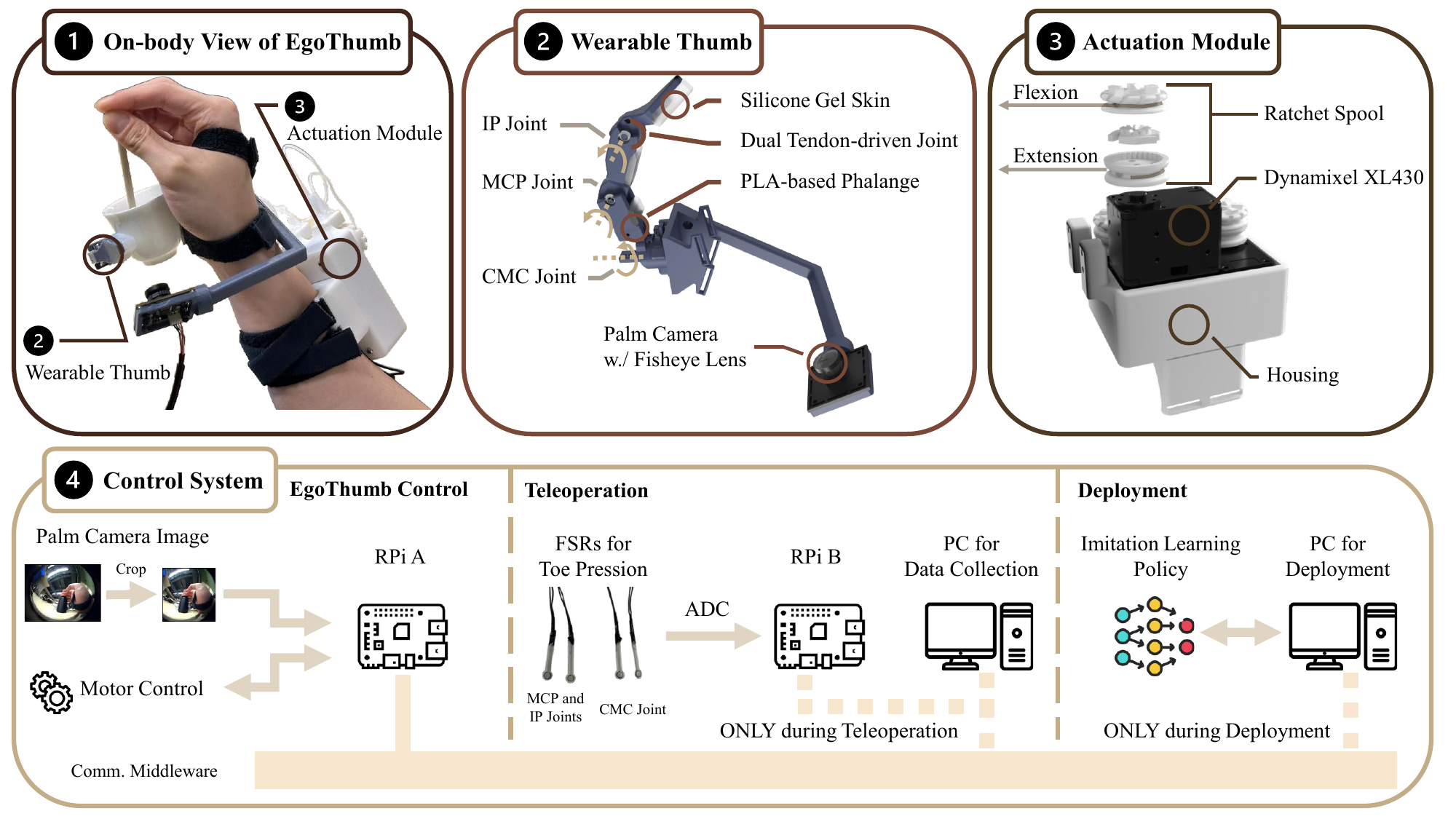}
  \caption{
  \textbf{Overview of the EgoThumb System Design.}
  \textbf{1) On-body View}: The system comprises a wearable thumb and a forearm-mounted actuation module.
  \textbf{2) Wearable Thumb}: The thumb features dual tendon-driven joints for precise control and PLA phalanges with a silicone-gel coating for stable grasping. A palm-facing fisheye camera provides egocentric visual feedback.
  \textbf{3) Actuation Module}: The actuation module houses three Dynamixel XL430-driven ratchet spools. This forearm-mounted configuration significantly reduces distal weight.
  \textbf{4) Control System}: Our EgoThumb interfaces with teleoperation and deployment systems via custom communication middleware.
  }
  \label{fig:system-design}

  \vspace{-1.5em} 
\end{figure*}

In this section, we present the architectural design of the EgoThumb system, the teleoperation framework utilized for data collection, and the formulation of the vision-driven imitation learning policy. 
To facilitate low-latency, cross-platform communication between these subsystems, we developed a communication middleware leveraging \verb|Zenoh| \cite{corsaro2023zenoh} and \verb|Protocol Buffers| \cite{protobuf}.
We use Network Time Protocol (NTP) \cite{mills2010network} to synchronize all the Raspberry Pis and PCs involved in the system.

\subsection{EgoThumb Design}

The EgoThumb system comprises two primary components: 1) a compact, tendon-driven wearable thumb with a wide field-of-view (FOV) palm-facing camera; and 2) a forearm-mounted actuation module.

\subsubsection{Wearable Thumb and Palm-Facing Camera}
The wearable thumb's kinematic structure mimics human anatomy (Fig. \ref{fig:system-design}.2), incorporating three joints: the carpometacarpal (CMC), metacarpophalangeal (MCP), and interphalangeal (IP). 
The MCP and IP joints are modeled as 1-degree-of-freedom (DoF) revolute joints, consistent with the biological thumb. 
While the biological CMC joint exhibits complex, multi-DoF kinematics \cite{kalshoven2024passive, halilaj2014vivo}, our system focuses primarily on grasping and holding tasks, making a high-DoF design unnecessary. 
Consequently, we simplify the CMC into a single-DoF revolute joint that rotates perpendicular to the palm plane \cite{kieliba2021robotic, clode2024evaluating}. 
To maintain terminological consistency across all of the device's joints, we refer to the CMC joint's abduction and adduction as flexion and extension throughout this paper.

To avoid the bulk associated with direct-drive mechanisms, we adopt a tendon-driven architecture inspired by recent anthropomorphic robotic hands \cite{christoph2025orca, zorin2025ruka, shadowrobotShadowRobot, dexhandDexHandOpen}. 
Specifically, we leverage the dual-tendon actuation design proposed by \cite{christoph2025orca}, where each joint is controlled by two tendons: one for flexion and one for extension. 
This dual-tendon mechanism enables active, precise position control in both directions. 
This stands in contrast to conventional tendon-driven SR fingers, which often rely on passive spring-based mechanisms for extension, leading to reduced control authority \cite{wu2015hold, weijun2024kinematic, kieliba2021robotic, clode2024evaluating}. 
The mechanical rotation ranges are $180^\circ$ for the CMC joint, $84.14^\circ$ for the MCP joint, and $141.34^\circ$ for the IP joint.
These ranges are determined by the joints' mechanical design and are sufficient to cover the grasping and holding postures required by our tasks.
A calibration procedure is conducted to map the physical joint positions to the corresponding motor angles in the actuation module \cite{christoph2025orca}.

The phalanges are fabricated using Fused Deposition Modeling (FDM) with Polylactic Acid (PLA). 
To enhance grasp stability and friction, a custom-molded layer of Shore 10A silicone (\verb|Dragon Skin 10|) is bonded to the palmar surface.
The device is secured to the user's hand via two adjustable hook-and-loop straps—one traversing the purlicue and the other encircling the wrist—ensuring a secure fit across varying hand sizes and preventing slippage during operation.

The system perceives hand motion and environmental context via a global-shutter camera (Fig. \ref{fig:system-design}.2) that is fixed to the wearable thumb on the palmar side of the hand and oriented from the hand root toward the fingers (i.e., palm-facing).
Equipped with a fisheye lens ($220^\circ$ FOV) \cite{tao2025dexwild}, the camera captures comprehensive visual data encompassing the hand, the robotic thumb, and the target object. 
Unlike DexUMI \cite{xu2025dexumi}, which mounts the camera on the ulnar side, we position our sensor on the radial side to maintain an unobstructed view of the interaction zone.
We favor this egocentric, palm-facing placement over a third-person camera to avoid the occlusions that arise as the wearable finger and hand move through the workspace.
The raw $240 \times 320$ RGB frames are asymmetrically cropped to $224 \times 224$ pixels, prioritizing the bottom-right quadrant. 
This cropping strategy serves as a spatial prior, centering the human-robot interaction workspace while filtering out extraneous background noise.
The policy consumes these raw cropped frames without geometric undistortion \cite{chi2024universal, xu2025dexumi} and requires no intermediate hand-pose estimation, avoiding on-device vision overhead.

\subsubsection{Actuation Module}
The actuation module (Fig. \ref{fig:system-design}.3) houses three actuator-spool units, each comprising a ratchet spool driven by a \verb|Dynamixel XL430-W250-T| servo motor. 
The ratchet spool anchors the two tendons, converting motor rotation into flexion/extension movements. 
A key feature of this design is the ability to rapidly re-tension the tendons to mitigate slack accumulation \cite{christoph2025orca}. 
The \verb|Dynamixel| motors provide a stall torque of 1.4 N$\cdot$m, ensuring stable grasping force, while their 12-bit encoders ($0.088^\circ$ resolution) enable precise proprioceptive feedback.

Tendons are routed from the wearable unit to the actuation module through low-friction Polytetrafluoroethylene (PTFE) tubes. 
The high axial stiffness of the PTFE tubing maintains invariant tendon path lengths during wrist articulation. 
This effectively decouples the robotic finger’s motion from the user’s wrist movements, ensuring consistent control regardless of hand pose.

A \verb|Raspberry Pi 3B+| (RPi A) serves as the embedded controller (Fig. \ref{fig:system-design}.4, Left), managing motor control loops at 60 Hz and image acquisition at 20 Hz.
It interfaces with the teleoperation and deployment systems via the aforementioned middleware.
The RPi runs only these control and acquisition loops; policy inference is offloaded to the external PC, so it performs no vision processing.
The 20 Hz acquisition rate is sufficient for the low-speed dynamics of our hold-and-manipulate tasks.

\subsection{Teleoperation System}
\label{sec:teleop-sys}

To facilitate data collection for imitation learning, we designed a customized foot-operated teleoperation interface \cite{kieliba2021robotic, clode2024evaluating} (Fig. \ref{fig:system-design}.4, Middle).
The system utilizes four Force Sensitive Resistors (FSRs), each integrated into a voltage divider with a $10k\Omega$ resistor and sampled by a 16-bit ADC (\verb|TI ADS1115|) at 50 Hz. 
Signal processing is handled by a secondary \verb|Raspberry Pi 3B+| (RPi B).

The control mapping is intuitive: the user modulates the left two FSRs with their left toe to control the flexion/extension velocity of the MCP and IP joints. 
Simultaneously, the right toe presses the right two FSRs to control the CMC joint. 
To reduce the control dimensionality for the operator, we enforce a linear kinematic synergy between the MCP and IP joints, defined as $\theta_{\text{IP}} = \frac{2}{3}\theta_{\text{MCP}}$.

During data collection, the palm-facing camera images, joint positions of the wearable thumb, and target actions from teleoperation are captured by RPi B and logged to a PC.
The teleoperation hardware is disconnected during autonomous deployment.

\subsection{Imitation Policy Learning}

We formulate the control of the EgoThumb as a vision-based imitation learning problem.
We adopt the widely used Diffusion Policy framework \cite{chi2024universal, chi2025diffusion}, which models multimodal action distributions and generates smooth action sequences suited to continuous joint control; the control policy is defined as $a_{t+1:t+H} = \pi(I_{t-T:t}, q_{t-T:t})$.
At time step $t$, $a_t$ represents the target action, $I_t$ denotes the cropped palm-facing camera image, and $q_t$ represents the current proprioceptive joint positions.
Following the short observation history of UMI \cite{chi2024universal}, we set $T=2$ and use a prediction horizon of $H=8$, which we found to work well on our tasks.

The visual input $I_t$ is processed by a CLIP-pretrained \cite{radford2021learning} Vision Transformer (ViT) \cite{dosovitskiy2020image} to extract robust latent representations, which are then concatenated with the proprioceptive states.
The core generative backbone is a ConditionalUnet1D, which maps the concatenated features to a sequence of future actions through iterative denoising.
To ensure precise synchronization between the asynchronous visual stream and hardware states, we implement continuous latency matching \cite{chi2024universal}. 
During deployment, as shown in Fig. \ref{fig:system-design}.4 (Right), the policy runs on an external PC, communicating real-time inference commands to the wearable system via the middleware.

\section{Experiments}
\label{sec:experiments}
Autonomous supernumerary fingers must dynamically adapt to human hand motions, accommodate various hand shapes, and function reliably across diverse environments.
To evaluate our proposed system, we investigate the following research questions:
\textbf{Q1}: Can our EgoThumb assist grasping and free human hands for other motion?
\textbf{Q2}: Can our system generalize to \textit{different hands}?
\textbf{Q3}: Can our system generalize to \textit{``in-the-wild'' environments}?

\begin{figure*}[!t]
  \centering
  \includegraphics[
    width=\textwidth,
    trim=0pt 0pt 0pt 0pt,
    clip
  ]{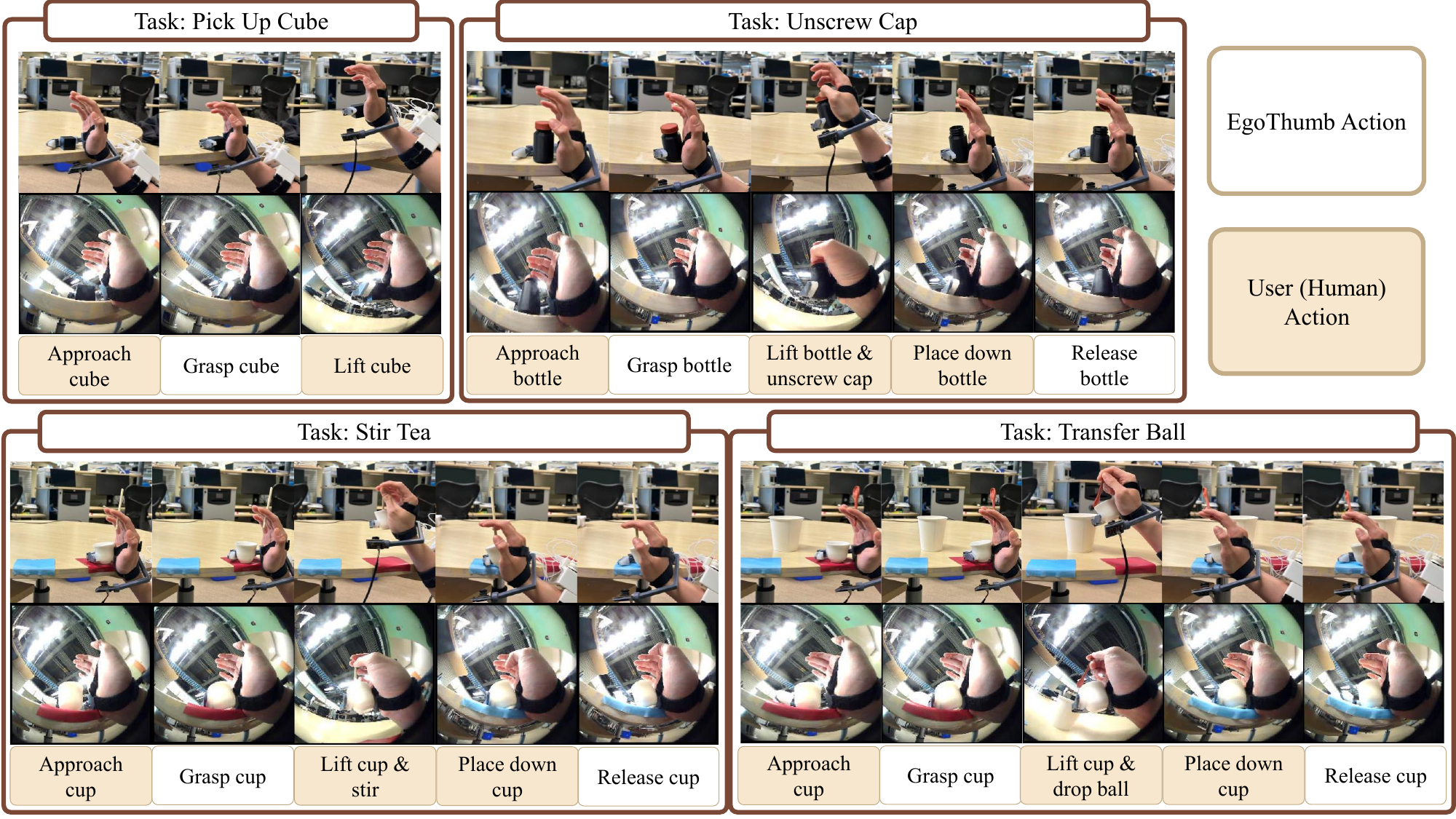}
  \caption{\textbf{Execution Sequences for the Four Evaluation Tasks.} Policy rollouts demonstrating collaborative human-robot manipulation across four tasks: \textbf{Pick Up Cube}, \textbf{Unscrew Cap}, \textbf{Stir Tea}, \textbf{Transfer Ball}. For each task, the top row displays the user's first-person view of the interaction, while the bottom row shows the corresponding egocentric palm-facing camera view used by the policy. The timelines below outline the coordinated sequence of autonomous EgoThumb actions and human motions.}
  \label{fig:tasks}
  \vspace{-1em} %
\end{figure*}

\subsection{Evaluation Setup}
We evaluate our wearable finger across four different real-world tasks that involve human-robot collaboration, adapted from prior hand-augmentation studies \cite{kieliba2021robotic, clode2024evaluating}, as shown in Fig. \ref{fig:tasks}:
\begin{itemize}
    \item \textbf{Pick Up Cube}: The user reaches toward a rigid cube placed in the initial area with their right hand. 
    The wearable thumb grasps the cube and then the user lifts the cube off the table. 
    \item \textbf{Unscrew Cap}: The user approaches a capped bottle in an initial area.
    The wearable thumb secures the bottle's body, enabling the user to lift it. While the device maintains a stable grasp, the user employs their biological fingers to pinch and rotate the cap. Finally, the user returns the bottle, and the wearable thumb disengages. 
    \item \textbf{Stir Tea}: 
    The user, holding a stirring stick, approaches a cup positioned in a red initial area. The wearable thumb secures and stabilizes the cup, allowing the user to lift it and stir inside. The user then transports the cup to a target blue area and places it down, at which point the wearable thumb disengages.
    \item \textbf{Transfer Ball}: A teacup containing a small ball is positioned at a red initial area. The user approaches the cup with a spoon. The wearable thumb secures the cup, allowing the user to lift and tilt it. The user then employs the spoon to guide the ball out of the tilted teacup and into a secondary target receptacle. Following the transfer, the user transports the teacup to a blue target region and places it down, prompting the wearable thumb to disengage.
\end{itemize}

A demonstrator collects 50 episodes of training dataset for each task using our teleoperation system (Sec. \ref{sec:teleop-sys}).
The objects are randomly placed at the initial area and moved to a random location of the target area for each task, as shown in Fig. \ref{fig:env} (Top, marked with darker regions). 
The imitation learning policy is trained on Intel I9-9960X and NVIDIA TITAN RTX for 120 epochs.

We conduct 10 evaluation episodes per task.
For each evaluation episode, we use the same initial and target areas as the training data.
To assess the generalization of our system to different hands, two other human operators (whose hands are not used in training data collection, User 1 and User 2) participate in this study.
We maintain a consistent initial object configuration across different users.
The evaluation is conducted on Intel I9-9960X and NVIDIA RTX 2080 Ti.
The policy rollouts for four tasks are shown in Fig. \ref{fig:tasks}.

\begin{table}[t]
\centering
\renewcommand{\arraystretch}{1.1}
\resizebox{\columnwidth}{!}{
\begin{tabular}{l|cccc|c}
\hline
 & \textbf{Pick Up Cube} & \textbf{Unscrew Cap} & \textbf{Stir Tea} & \textbf{Transfer Ball}
 & \textbf{Avg.} \\
\hline
\textbf{Demonstrator} & 1.0 & 0.7 & 0.9 & 1.0 & 0.90 \\
\textbf{User 1}       & 1.0 & 0.9 & 0.8 & 1.0 & 0.925 \\
\textbf{User 2}       & 1.0 & 0.8 & 0.8 & 0.9 & 0.875 \\
\hline
\end{tabular}
}
\caption{
\textbf{Success Rates across Different Hands.}
We report the performance when objects are placed within initial and target areas in the training data (Fig.~\ref{fig:env}, Top, darker regions).
EgoThumb system achieves strong performance across all four tasks when operated by the demonstrator, and it generalizes effectively to novel participants (User 1 and User 2).
}
\label{tab:task_success_rates}
\end{table}

\begin{table}[t]
\centering
\renewcommand{\arraystretch}{1.1}
\resizebox{\columnwidth}{!}{
\begin{tabular}{l|cccc}
\hline
 & \textbf{Pick Up Cube} & \textbf{Unscrew Cap} & \textbf{Stir Tea} & \textbf{Transfer Ball} \\
\hline
\textbf{Demonstrator} & 1.0 & 0.8 & 0.6 & 0.2 \\
\textbf{User 1}       & 1.0 & 0.7 & 0.3 & 0.1 \\
\textbf{User 2}       & 1.0 & 0.7 & 0.7 & 0.2 \\
\hline
\textbf{Avg.} & 1.00 & 0.73 & 0.53 & 0.17 \\
\hline
\end{tabular}
}
\caption{
\textbf{Success Rates in Expanded Initial and Target Areas.}
We evaluate our system on an expanded initial and target areas (Fig.~\ref{fig:env}, Top, lighter regions) compared to the training data. Results show strong robustness on \textbf{Pick Up Cube} and \textbf{Unscrew Cap}, moderate robustness on \textbf{Stir Tea}, and a clear drop on \textbf{Transfer Ball}, likely due to increased foreground visual variation and its longer-horizon manipulation requirements.
}
\label{tab:task_success_rates_L}
\vspace{-2em}
\end{table}

\subsection{Results and Analysis}

\textbf{Our wearable finger enables the human hand to perform concurrent tasks by providing stable grasp assistance (Q1)}: 
As detailed in Tab. \ref{tab:task_success_rates} (Demonstrator row), our system achieves an average success rate of 0.9 across all four tasks when evaluated on the demonstrator's hand (the in-distribution hand). 
In every scenario, the system demonstrates the ability to: 1) precisely determine the optimal timing for grasping and releasing objects; and 2) maintain a stable grasp during dynamic hand interactions.
We attribute this to our occlusion-free camera placement and the dual-tendon, silicone-skinned finger that ensures secure grasping and efficient release.

\textbf{Our finger system demonstrates robustness across varying hand morphologies (Q2)}:
Adjustable hook-and-loop fasteners allow the wearable thumb to physically adapt across users. However, novel hands introduce out-of-distribution (OOD) visual scenarios, since training data was collected solely from the demonstrator’s hand.
Specifically, anatomical variations alter the human hand's appearance and relative pose within the camera's field of view. 
Despite these visual discrepancies, Tab. \ref{tab:task_success_rates} (User 1 and User 2 rows) illustrates that our system maintains average success rates at least 0.875 for both User 1 and User 2 across all four tasks.
This sustained success suggests strong generalization to novel users; the system effectively learns to attend to human-object interactions and accurately identifies the optimal timing and regions for grasping. 
We attribute this robustness to the rigidly mounted camera, which keeps the robotic thumb's pose constant in the visual frame regardless of hand size while centering the view on the interaction and its environmental context.

\begin{figure}[!t]
  \centering
    \includegraphics[
    width=0.47\textwidth,
    trim=49pt 30pt 90pt 255pt,
    clip
    ]{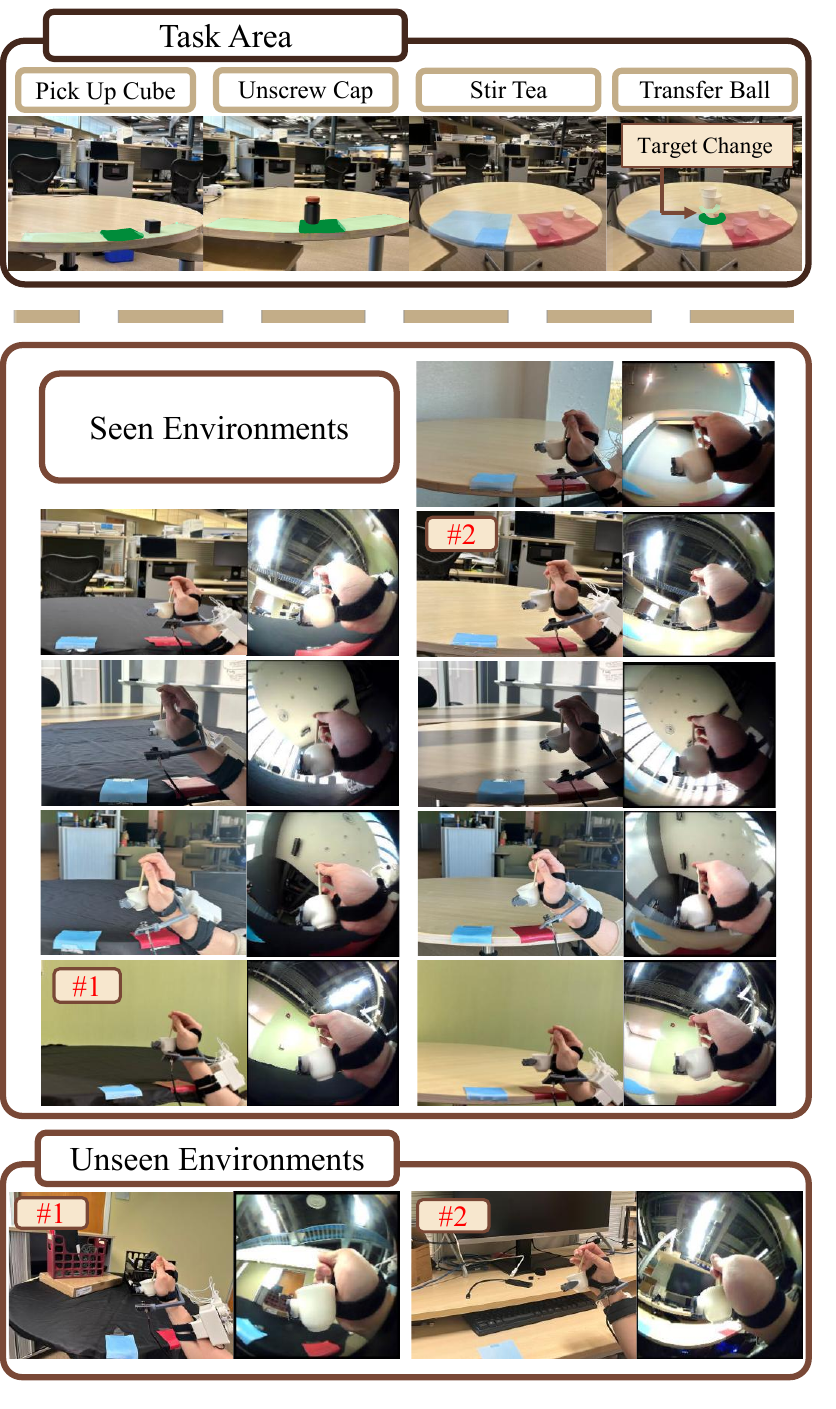}
  \caption{
  \textbf{Task Area and ``In-the-Wild'' Environments.} 
  \textbf{Top (Task Area):} Darker areas represent the initial and target areas for training data, while the larger, lighter areas (including the original darker areas) represent the expanded areas used for the robustness test of object distributions. \textbf{Pick Up Cube} and \textbf{Unscrew Cap} utilize the green areas, expanded from 10 cm $\times$ 10 cm to 50 cm $\times$ 10 cm. \textbf{Transfer Ball} and \textbf{Stir Tea} use red for pick-up and blue for drop-off, expanded from 10 cm $\times$ 10 cm to 30 cm $\times$ 25 cm. For the Transfer Ball task, expanding the evaluation area shifts the target receptacle's position (``Target Change''). \textbf{Bottom (``In-the-wild'' Environments):} The ``Seen Environments'' panel displays the diverse environments included in the broadened training dataset. The ``Unseen Environments'' panel displays completely novel backgrounds used strictly to test out-of-distribution generalization. Labels \#1 and \#2 denote the specific seen and unseen test environments.}
  \label{fig:env}
  \vspace{-2em} %
\end{figure}

\textbf{Our system is robust to variations in object distributions (Q3)}:
To assess resilience against spatial variations, we evaluated the system by enlarging the initial and target areas for the objects compared to training data, as depicted in Fig. \ref{fig:env} (Top, lighter regions).
As shown in Tab. \ref{tab:task_success_rates_L}, the system consistently achieves average success rates over 0.73 in the \textbf{Pick Up Cube} and \textbf{Unscrew Cap} tasks across all users.
Performance degrades slightly to an average success rate of 0.53 for the \textbf{Stir Tea} task and drops to 0.17 for the \textbf{Transfer Ball} task.
We hypothesize that these variances stem from differing visual and temporal complexities.
The \textbf{Pick Up Cube} task serves as a robust baseline despite spatial variation.
Similarly, in the \textbf{Unscrew Cap} task, although the initialization area is expanded, the primary background remains static and distant, preserving critical visual features. 
In contrast, the \textbf{Stir Tea} task suffers because variations in the initial and target zones introduce significant foreground color and texture shifts, creating substantial visual distractors.
Finally, the \textbf{Transfer Ball} task exhibits the lowest performance due to its long-horizon nature, which demands precise sequential planning for identifying grasp regions, stabilizing the object during interaction, and executing timely releases.

\begin{table}[t]
\scriptsize
\centering
\setlength{\tabcolsep}{8pt}
\renewcommand{\arraystretch}{1.2}

\begin{tabular}{l|cc|cc}
\hline
 \textbf{Envs} & \textbf{Seen \#1} & \textbf{Seen \#2} 
 & \textbf{Unseen \#1} & \textbf{Unseen \#2} \\
\hline

\textbf{Demonstrator} & 0.8 & 1.0 & 0.7 & 0.9 \\
\textbf{User 1}       & 0.9 & 1.0 & 0.7 & 0.6 \\
\textbf{User 2}       & 0.7 & 1.0 & 0.6 & 0.7 \\

\hline
\textbf{Avg. User}      & 0.80 & 1.00 & 0.67 & 0.73 \\
\hline
\end{tabular}

\caption{\textbf{Task Success Rates Across Different Backgrounds.} 
We evaluate our policy on two seen and two unseen environments using the Stir Tea Task.
The policy maintains robust performance across both seen and entirely novel (unseen) backgrounds for all users.}
\label{tab:bg_success}
\vspace{-3em} %
\end{table}

\begin{figure}[ht]
  \centering
  \includegraphics[
    width=\columnwidth
  ]{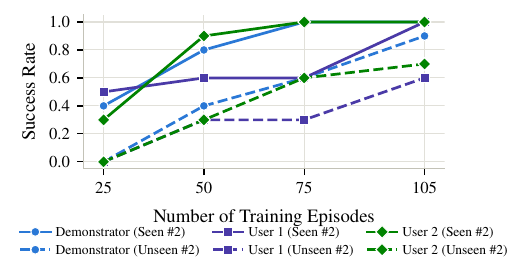}
  \caption{\textbf{Impact of Training Data Scale on Generalization.} 
  Scaling the demonstrations from 25 to 105 generally improves policy performance in both seen (solid lines) and unseen (dashed lines) environments across all users, proving that more data drives robust out-of-distribution generalization.}
  \label{fig:diff-train-size}
\end{figure}

\textbf{Our finger system demonstrates robustness to background variations (Q3)}:
To evaluate robustness against diverse visual backgrounds and lighting conditions, we conducted an extended ``in-the-wild'' study on the \textbf{Stir Tea} task. 
The control policy was trained on a broadened dataset of 105 demonstrations collected across 9 distinct environmental settings (Fig. \ref{fig:env}, Seen Environments). 
We evaluated the policy in two in-distribution (seen) environments and two completely novel (unseen) environments (marked \#1 and \#2 in Fig. \ref{fig:env}, Bottom).
Tab. \ref{tab:bg_success} shows that our system maintains average success rates above 0.8 across three users in the seen environments.
In unseen environments, performance declines slightly but remains above 0.67.
To investigate the drivers of this generalization, we varied the training dataset size and evaluated performance on environments Seen \#2 and Unseen \#2.
As illustrated in Fig. \ref{fig:diff-train-size}, scaling the number of training demonstrations yields a corresponding increase in success rates across both settings for all users.
This confirms that the system effectively learns generalizable visual representations invariant to environmental distractors, demonstrating that data scaling successfully drives out-of-distribution robustness rather than simply overfitting to specific backgrounds.

\subsection{Ablation Study}

We conducted an ablation study on the latency matching and image cropping modules to isolate their contributions. Both ablations were trained and evaluated on the \textbf{Unscrew Cap} task using the in-distribution demonstrator. To ensure a fair comparison, the training dataset, policy architecture, and hyperparameters remained identical to the main experiments. Quantitative results are detailed in Tab. \ref{tab:ablation_results}.

\begin{table}[t]
\centering
\setlength{\tabcolsep}{12pt}
\renewcommand{\arraystretch}{1.15}
\begin{tabular}{lc}
\hline
\textbf{System Configuration} & \textbf{Success Rate} \\
\hline
\textbf{Full System } & 0.7 \\
\textbf{w/o Latency Matching} & 0.0 \\
\textbf{w/o Image Cropping} & 0.0 \\
\hline
\end{tabular}
\caption{\textbf{Ablation Study Results.} 
Disabling either latency matching or image cropping results in catastrophic policy failure, confirming their necessity for robust visuo-motor control.}
\label{tab:ablation_results}
\scriptsize
\vspace{-2em}
\end{table}
\noindent\textit{1) \textbf{Latency Matching:}}
Precise temporal alignment is critical for continuous control; latency matching synchronizes multimodal observation streams and mitigates hardware pipeline delays \cite{chi2024universal}. 

\textbf{Ablated Setup:} 
We disabled all temporal compensation. During training, we replaced latency-compensated continuous interpolation with a naive nearest-neighbor approach, matching images to the chronologically closest joint state. During inference, we queried the most recent joint position instead of interpolating to the image timestamp. Additionally, we disabled the forward predictive action delay—which normally offsets the computational latency between policy querying and physical execution—and executed predictions immediately.

\textbf{Results:} 
The ablated policy failed completely (0 success rate vs. 0.7 for the full system). This catastrophic drop confirms that temporal alignment and pipeline delay compensation are necessary for dynamic, high-precision human-robot coordination.

\noindent\textit{2) \textbf{Image Cropping:}}
Explicit spatial priors are crucial for filtering spurious environmental distractors and learning robust visuo-motor policies.

\textbf{Ablated Setup:} 
We bypassed our standard image cropping—which normally isolates the primary region of interest (the hand and target object), as shown in Fig. \ref{fig:system-design}.4 (Left). 
Instead, uncropped camera frames were fed directly into an ImageNet-pretrained \cite{deng2009imagenet} ResNet34 \cite{he2016deep} encoder during both training and inference. 
We use ResNet34 because it can directly process the uncropped 240 $\times$ 320 input resolution, whereas CLIP does not support this input size.

\textbf{Results:} %
The policy with uncropped image input also failed completely (0 success rate vs. 0.7 for the full system). Without this spatial bottleneck, the policy cannot isolate fine-grained manipulation features and likely overfits to background noise, proving that image cropping is an indispensable preprocessing step for our vision pipeline.

\section{Conclusion and Limitation}
\label{sec:conclusions}
We presented EgoThumb, an open-source, vision-driven supernumerary thumb that augments human dexterity in ``hold-and-manipulate'' tasks.
By pairing a lightweight, compact, tendon-driven joint design with egocentric visual imitation learning, the system autonomously decodes human intent and generalizes robustly across diverse users and environments, validating our palm-facing camera design for capturing invariant visual representations.
Several limitations remain and motivate future work.
Performance degrades on long-horizon tasks with large spatial variation in the initial and target areas (\textbf{Stir Tea} and \textbf{Transfer Ball}), motivating hierarchical learning frameworks.
Rapid hand motions can induce motion blur; this does not affect our low-speed hold-and-manipulate tasks, but high-speed manipulation remains unevaluated.
The palm-facing camera protrudes from the hand and may reduce wearing comfort, motivating a more compact, integrated sensor.
Our data were collected under relatively consistent hand approach angles, leaving robustness to varied approach angles untested.
Finally, our evaluation relies on binary task success over 10 trials per condition; finer measures of coordination quality (e.g., grasp timing and motion smoothness) and larger-scale evaluation with statistical analysis are also needed.

\printbibliography

\end{document}